\documentclass[10pt,journal,compsoc]{IEEEtran}
\usepackage{multirow}
\usepackage{epsfig}
\usepackage{graphicx}
\usepackage{amsmath}
\usepackage{amssymb}

\usepackage[colorlinks,linkcolor=blue]{hyperref}
\usepackage{url}
\usepackage{booktabs}

\ifCLASSOPTIONcompsoc
  \usepackage[nocompress]{cite}
\else
  \usepackage{cite}
\fi
\usepackage{algorithm} 
\usepackage{algorithmic}

\usepackage[english]{babel}

\newlength\savewidth
\newcommand\shline{\noalign{\global\savewidth\arrayrulewidth
                           \global\arrayrulewidth 1.0pt}%
                  \hline
                  \noalign{\global\arrayrulewidth\savewidth}}

\begin{document}
\title{Semi-Supervised Domain Generalizable \\Person Re-Identification}

\author{Lingxiao He, Wu Liu, Jian Liang, Kecheng Zheng, Xingyu Liao, Peng Cheng, Tao Mei,~\IEEEmembership{IEEE Fellow}
\IEEEcompsocitemizethanks{\IEEEcompsocthanksitem Lingxiao He, Wu Liu, Xingyu Liao, Peng Cheng and Tao Mei are with JD AI Research, Beijing, 100101, China. Corresponding authors: Lingxiao He (e-mail: helingxiao3@jd.com) and Wu Liu (e-mail:liuwu1@jd.com) \protect 

\IEEEcompsocthanksitem Jian Liang is with Institute of Automation, Chinese Academy of Sciences (CASIA), Beijing, 100101, China.

\IEEEcompsocthanksitem Kecheng Zheng is with Institute of Automation, University of Science and Technology of China (USTC), Heifei, 230026, China. 

}
\thanks{}}

\markboth{Journal of \LaTeX\ Class Files,~Vol.~14, No.~8, August~2015}%
{Shell \MakeLowercase{\textit{et al.}}: Bare Demo of IEEEtran.cls for Computer Society Journals}

\IEEEtitleabstractindextext{%
\begin{abstract}
Despite the success in cross-camera person matching, existing person re-identification (re-id) methods are stuck when deployed to a new unseen scenario.
Recent efforts have been substantially devoted to domain adaptive person re-id where extensive unlabeled data in the new scenario are utilized in a transductive learning manner. 
However, for each scenario, it is required to first collect enough data and then train such a domain adaptive re-id model, thus restricting their practical application.
Instead, we aim to explore multiple labeled datasets to learn generalized domain-invariant representations for person re-id, which is expected universally effective for each new-coming re-id scenario. 
To pursue practicability in real-world systems, we collect all the person re-id datasets (20 datasets) in this field and select the three most frequently used datasets (i.e., Market1501, DukeMTMC, and MSMT17) as unseen target domains.
In addition, we develop DataHunter that collects over 300K+ weak annotated images named YouTube-Human from YouTube street-view videos, which joins 17 remaining full labeled datasets to form multiple source domains.
On such a large and challenging benchmark called \textbf{FastHuman} ($\sim$ 440K+ labeled images), we further propose a simple yet effective Semi-Supervised Knowledge Distillation (SSKD) framework. 
SSKD effectively exploits the weakly annotated data by assigning soft pseudo labels to YouTube-Human for improving the generalization ability of models. 
Experiments on several protocols verify the effectiveness of the proposed SSKD framework on domain generalizable person re-id, which is even comparable to supervised learning on the target domains. 
Lastly, but most importantly, we hope the proposed benchmark \textbf{FastHuman} could bring the next development of domain generalizable person re-id algorithms. 
Code and datasets will be available at \url{https://github.com/JDAI-CV/fast-reid/projects/DG-ReID}.
\end{abstract}

}

\maketitle

\IEEEdisplaynontitleabstractindextext

\IEEEpeerreviewmaketitle

\IEEEraisesectionheading{\section{Introduction}\label{sec:introduction}}

\IEEEPARstart{P}erson Re-identification (re-id) \cite{he2020guided, he2019foreground, luo2019bag, he2020fastreid, zhang2020ordered} has achieved remarkable success under the transductive domain adaptive setting with the assumption that training and testing data are easily collected. 
However, this assumption is often violated in practical applications as new domain data are not always accessible due to privacy issues and expensive labeling costs, leading to the dramatically dropped performance on the unseen target domain. 
For example, the model built on the data collected in the daytime domain performs poorly on unseen nighttime domains.
Therefore, recent efforts have been devoted to the Domain Generalizable Re-IDentification (dubbed as DG-ReID) problem that aims to learn a general domain-agnostic model, which can generalize well to unseen target domains.
\begin{figure}[t]
\centering
\vspace{-1em}
\includegraphics[height=4.4cm]{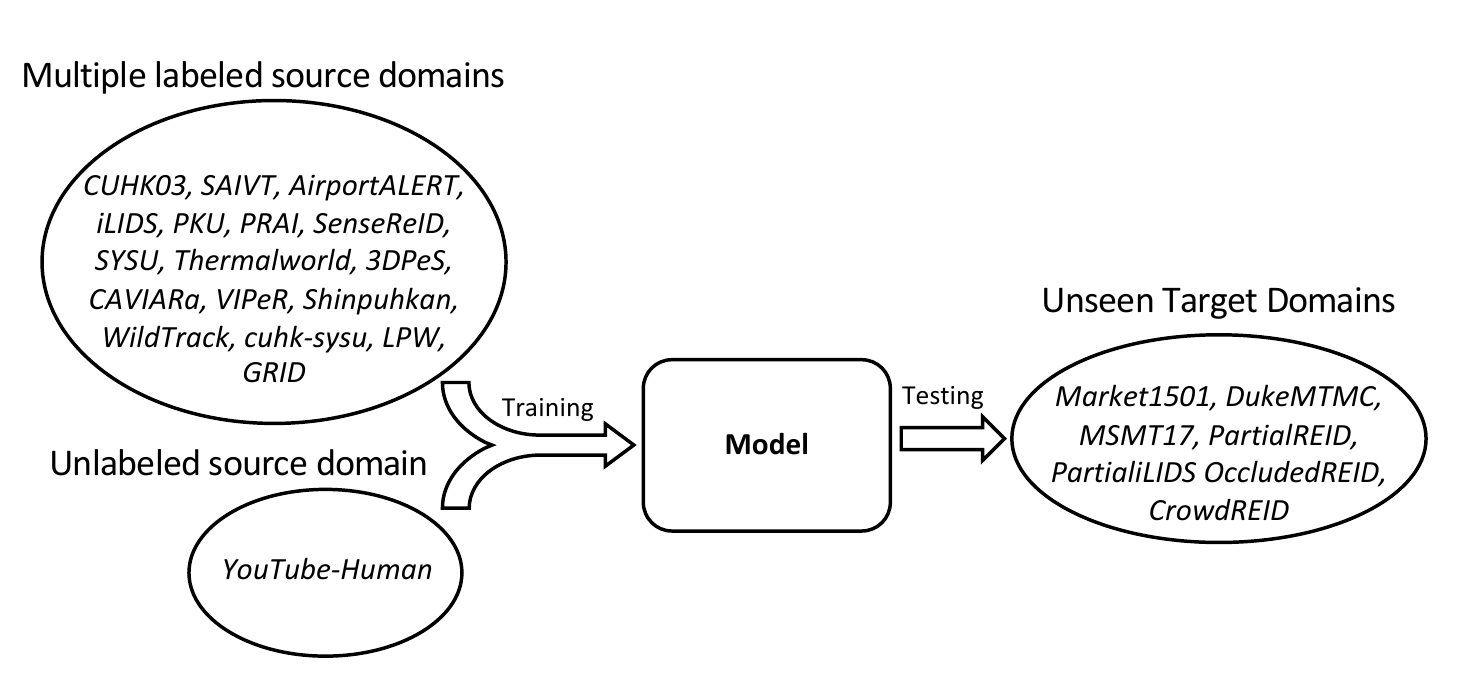}
\vspace{-2em}
\caption{Semi-supervised DG-ReID learns domain-agnostic features from multiple labeled source domains along with an unlabeled source domain, and the trained model can generalize well to unseen target domains when deploying to practical application scenarios.}
\vspace{-1em}
\label{fig1}
\end{figure}

To solve this challenging problem, data augmentation methods \cite{devries2017improved,yun2019cutmix,zhang2017mixup,matsuura2020domain} are developed to expand the source domain with diverse samples. 
Some other augmentation methods \cite{sun2019dissecting, wang2020surpassing} exploit the 3D body generation engine during data synthesis with different viewpoints, poses, illuminations, backgrounds, occlusions, and resolutions. 
However, it is difficult for these methods to simulate the realistic data distribution and may bring some noise interference that makes the learned model even worse. 
Besides, model-driven approaches \cite{choi2020meta, zhao2020learning} exploit meta-learning to simulate the train-test process of domain generalization, but they cannot well handle multiple source domains with limited fitting ability. 
On the contrary, we resort to consider unlabeled data from the web to increase the diversity of source data. 
In particular, we present an automatic data acquisition system to collect large-scale human images named YouTube-Human from the YouTube website, and it contains 300K+ weak annotation images captured by a single mobile camera from many street view videos. 

To address this new DG-ReID problem in Fig.~\ref{fig1}, this paper proposes a new framework called \textbf{S}emi-supervised \textbf{K}nowledge \textbf{D}istillation (\textbf{SSKD}), which fully exploits weak annotations in YouTube-Human to learn domain-general discriminative features. 
SSKD first trains a student model (e.g., ResNet-50) and a teacher model (e.g., ResNet-101) using labeled data from multi-source domain datasets. 
Then, SSKD develops an auxiliary classifier to imitate the soft predictions of unlabeled data generated by the teacher model. 
Meanwhile, the student model is also supervised by hard labels and predicted soft labels by the teacher model for labeled data. 

Concerning the testbed, we collect 20 domain publicly available datasets and merge them into a large-scale multi-source person re-id dataset named \textbf{FastHuman}.
\textbf{FastHuman} consists of 440K+ images from 38K+ subjects captured by 82 cameras and is collected from some realistic application scenarios such as airport, campus, street, and railway station, ensuring the diversity of the data.  
To evaluate the person re-id performance, three popular dataseets, i.e., Market1501 \cite{zheng2015scalable}, DukeMTMC \cite{zheng2017unlabeled}, and MSMT17 \cite{wei2018person}, are chosen as unseen domains for testing, and the remaining 17 domain datasets and YouTube-Human dataset are used as source domains for model training. 

To summarize, our contributions are as follows:
\begin{itemize}
    \item We present DataHunter that collects weakly annotated person images from YouTube to improve the generalization ability of DG-ReID.
    \item We propose a new semi-supervised DG-ReID framework that can effectively improve the generalization capability of person re-id models to unseen target domains by using labeled data and unlabeled data;
    \item We develop a large-scale benchmark named \textbf{FastHuman} to advance the development of DG-ReID task, which contains considerably diverse images from many scenarios and cameras. 
\end{itemize}


\section{Related Work}

\subsection{Domain Generalizable Re-IDentification} 
Generalization capability to unseen domains is crucial for person re-id models when deploying to practical applications. 
To address this problem, many data augmentation methods, e.g., random erasing \cite{zhong2020random}, Cutout \cite{devries2017improved}, CutMix \cite{yun2019cutmix}, and Mixup \cite{zhang2017mixup}, are widely used to increase the diversity of training data for improving the generalization of models.
Besides, virtual image generation methods \cite{sun2019dissecting, wang2020surpassing} also provide a feasible solution that uses the 3D body generation engine to generate diverse training data. 
However, it is difficult to simulate the real-world data and the improved performance is fairly limited due to some unwanted noises. 
In contrast to data-driven methods above, model-driven methods \cite{zhao2020learning,choi2020meta,li2019episodic,jin2020style} mainly exploit meta-learning for Domain Generalizable Re-IDentification (DG-ReID). 
Zhao \textit{et al.} \cite{zhao2020learning} propose the Memory-based Multi-Source Meta Learning (M3L) framework to train a generalizable model for unseen domains, in which a meta-learning strategy is introduced to simulate the train-test process of domain generalization. 
Choi \textit{et al.} \cite{choi2020meta} propose Meta Batch-Instance Normalization (MetaBIN) that generalizes normalization layers by simulating unsuccessful generalization scenarios beforehand in the meta-learning pipeline. 
Li \textit{et al.} \cite{li2019episodic} design an episodic training strategy that simulates train-test domain-shift using different domains during training to improve the robustness and generalization of the trained model to unseen domains.
Although these methods achieve satisfying results, they would become invalid when more datasets are involved in the source domain.

Recent years have witnessed the introduction of multiple person re-id datasets.
Li \textit{et al.} \cite{li2014deepreid} propose a challenging dataset called CUHK03, and
Zheng \textit{et al.} \cite{zheng2015scalable} propose a high-quality dataset named Market1501 for person re-id. 
Besides, other large-scale datasets like DukeMTMC \cite{zheng2017unlabeled} and MSMT17 \cite{wei2018person} are also the mainstream datasets for re-id evaluation.
Recently, almost all re-id approaches using single-domain (e.g., Market-1501 training, Market-1501 testing) or cross-domain (e.g., Market-1501 training, DukeMTMC testing) testing protocols for evaluating the re-id performance. 
However, person re-id is a problem oriented to practical applications varied with background, weather, seasons, illumination, occlusion, and other unpredictable factors. 
Only using a single-domain or cross-domain testing benchmark cannot verify the generalization ability of models, even used in practical application scenarios. 
To evaluate DG-ReID task, Zhao \textit{et al.} \cite{zhao2021learning} propose to combine multiple source datasets including Market1501, DukeMTMC, CUHK03 and MSMT17 datasets to train a model and use four small datasets including PRID~\cite{hirzer2011person}, GRID \cite{loy2013person}, VIPeR \cite{gray2008viewpoint} and iLIDS \cite{zheng2009associating} as testing datasets. 
Nevertheless, the number of images in the four testing sets is fairly small, which cannot effectively reflect the generalization ability of models. 
Besides, Dai \textit{et al.} \cite{dai2021generalizable} propose a new protocol of the leave-one-out setting for the existing large-scale public datasets such as Market1501, DukeMTMC, CUHK03 and MSMT17. Specifically, the leave-one-out setting selects one domain from these datasets for testing and all the remaining domains for model training (all images in these datasets). 
Compared to these existing benchmarks, the proposed benchmark has more diverse training samples and a more challenging testing set.

\begin{figure*}[t]
\centering
\includegraphics[height=6.4cm]{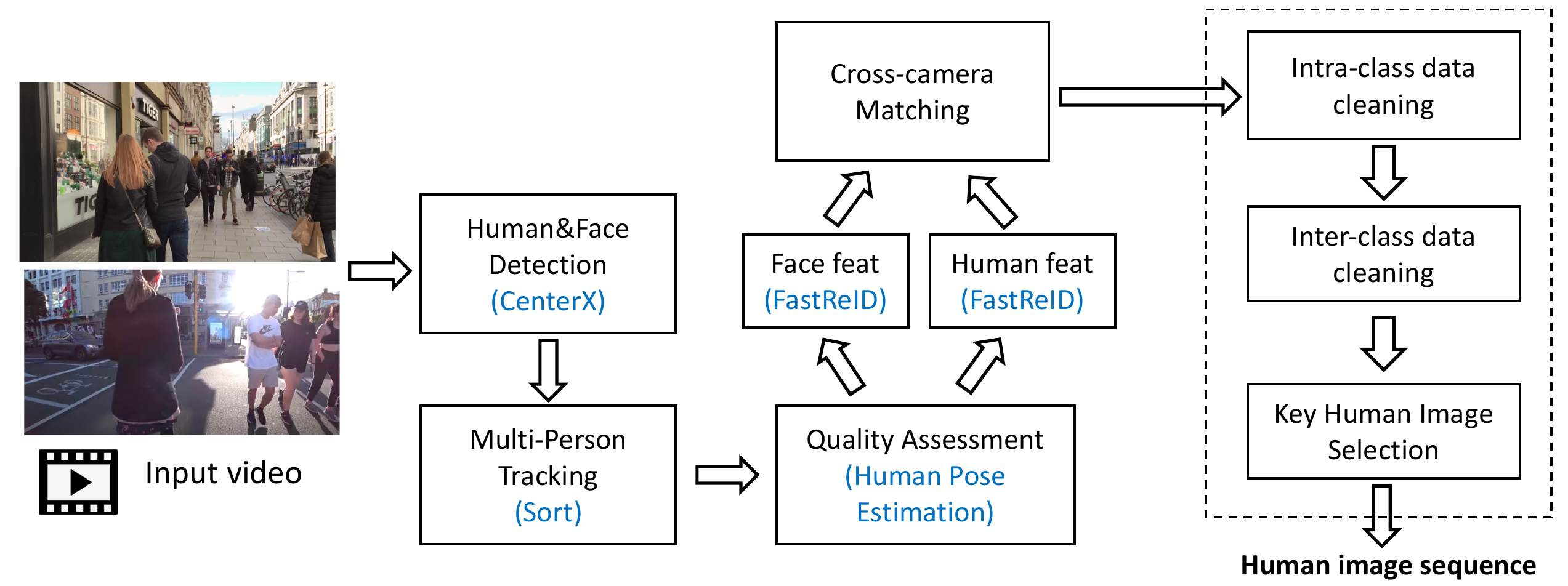}
\caption{The framework of DataHunter, uses automatic machine annotations to generate an amount of weakly annotated data to improve the generalization ability of models. 
DataHunter makes full use of face and human joint detection, pedestrian tracking, quality assessment, and machine learning-based intra-class and inter-class data cleaning modules to greatly improves the pre-labeling efficiency and accuracy. 
Benefiting from it, we are expected to save a lot of image labeling cost and provide high-quality data labeling.}
\label{fig2}
\end{figure*}

\subsection{Knowledge Distillation}
The learning of a small network from a large network is later formally popularized as vanilla knowledge distillation (KD) \cite{hinton2015distilling} where a small student model is generally supervised by a large teacher model.
The main idea is that the student model mimics the teacher model to achieve competitive or even superior performance. 
Particularly, knowledge is transferred from the teacher model to the student by minimizing a loss function in which the target is the distribution of class probabilities predicted by the teacher model. 

Existing knowledge distillation methods can be grouped into the three categories. 
The response-based Knowledge knowledge distillation uses the logits of a large deep model as the teacher knowledge~\cite{hinton2015distilling,mirzadeh2020improved,kim2018paraphrasing}. 
The activations, neurons or features of intermediate layers also can be adopted as the feature-based knowledge distillation~\cite{ahn2019variational,heo2019knowledge,komodakis2017paying} to instruct students in the learning of the model. 
The relationship-based knowledge distillation~\cite{lee2019graph,yu2019learning,liu2019knowledge,tung2019similarity} further explores the relationships between different activations, neurons or pairs of samples.
KD, as a quite popular technique, has been widely applied in model compression, object detection, and various feature learning tasks. 
Motivated by KD, we provide a feasible solution to combine labeled and unlabeled data to improve the generalization ability of person re-id models.


\subsection{Semi-Supervised Learning}
Semi-supervised learning (SSL) is a sophisticated field with a large number of methods developed in the literature. Existing SSL methods related to our work can be grouped into two main categories: consistency regularization~\cite{tarvainen2017mean,sajjadi2016regularization} and pseudo-labeling~\cite{xie2020self,berthelot2019mixmatch,sohn2020fixmatch}. 
As for consistency regularisation, the model~\cite{tarvainen2017mean,sajjadi2016regularization} predicts the labels of unlabelled samples, and these predictions should be consistent for perturbed versions of the same samples.
pseudo-labeling utilizes the predicted labels by a pre-trained model~\cite{xie2020self} or the model being trained~\cite{berthelot2019mixmatch,sohn2020fixmatch} for unlabeled samples with high confidence and use these predictions for model training.
Generally, semi-supervised Learning is relevant to semi-supervised domain generalization (SSDG) as both problems require the processing of unlabelled data. 
Nonetheless, it is more challenging to deal with the marginal data distribution of unlabelled data in SSDG as they are collected from heterogeneous domains.

\section{Domain Generalized Re-identification}
\subsection{DataHunter}
To acquire more cheap training data, we develop an automatic data labeling system named DataHunter. 
As shown in Fig.~\ref{fig2}, it integrates pedestrian detection in CenterX\footnote{\url{https://github.com/JDAI-CV/centerX}}, multi-object tracking, quality assessment algorithm and face recognition and person re-identification algorithms in FastReID\footnote{\url{https://github.com/JDAI-CV/fastreid}} to generate large-scale weakly annotated person images. 
Firstly, we detect multiple person bounding boxes, and then we employ the sorting algorithm to track multiple moving persons. 
To guarantee the quality of the collected human images, we use pose estimation to ensure the integrity of person images and confidence of keypoints for filtering out the poor quality images. 
Finally, we perform intra-class and inter-class data cleaning modules to reduce wrong labeling images.

\begin{figure}[t]
\centering
\includegraphics[height=3cm]{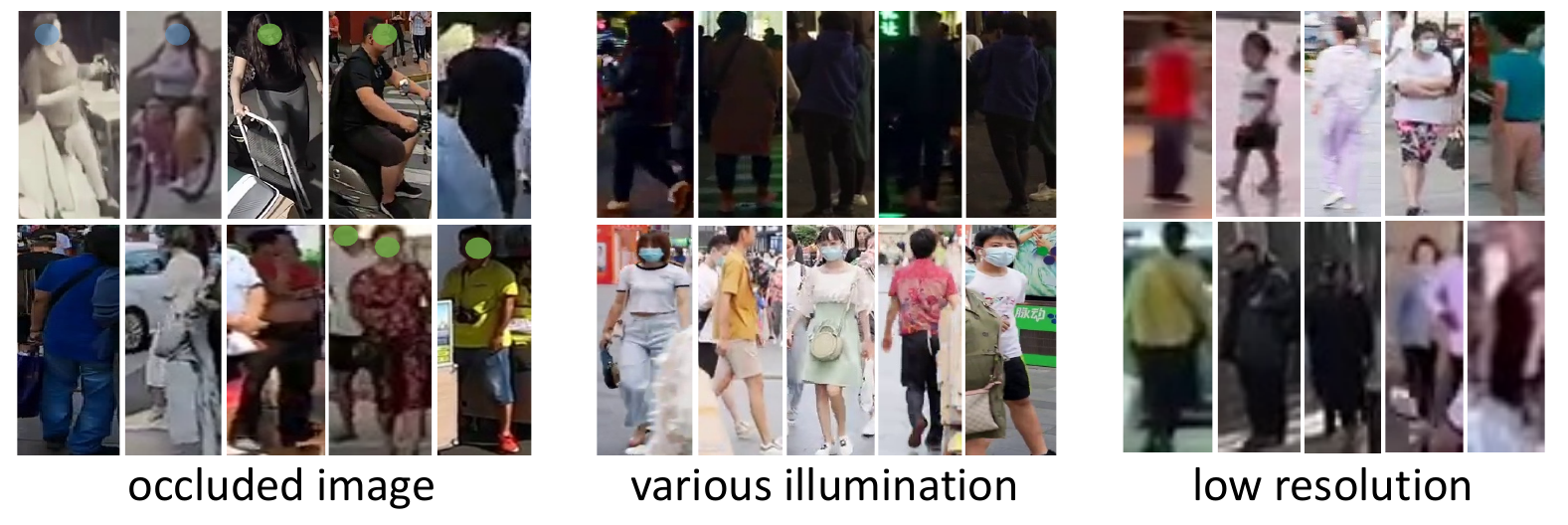}
\vspace{-2em}
\caption{Some examples of YouTube-Human. YouTube-Human contains some occluded images. Besides, images are captured with different light intensity and resolution.}
\label{fig3}
\end{figure}
\begin{figure*}[t]
\centering
\includegraphics[height=5cm]{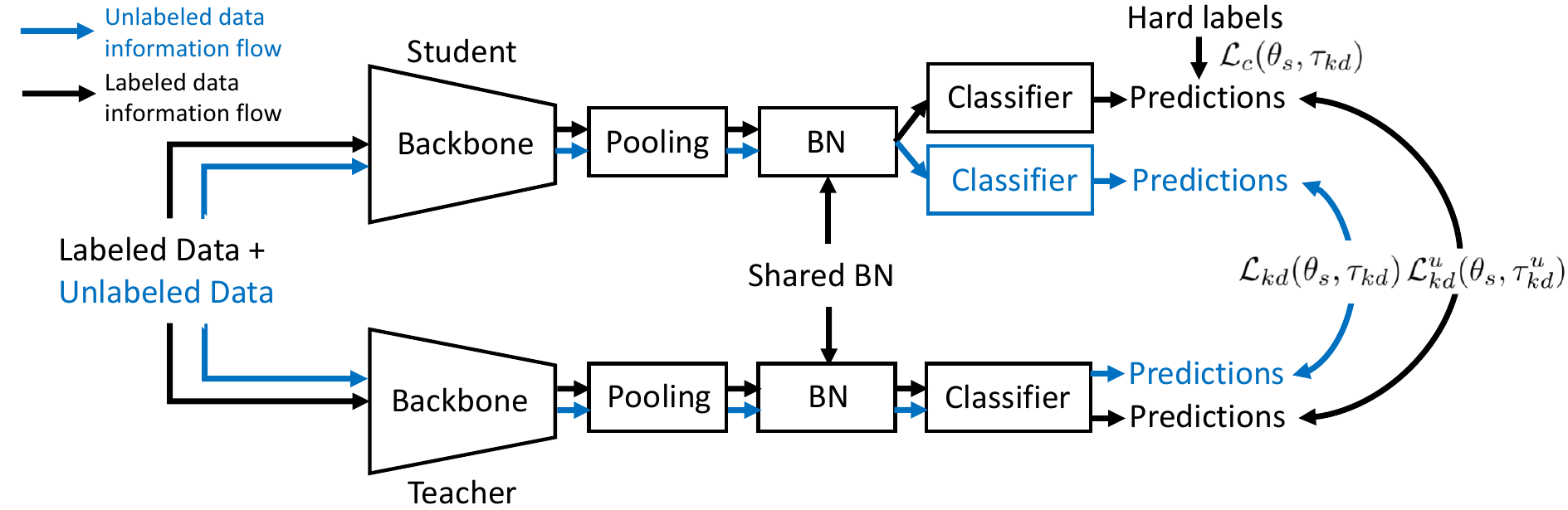}
\caption{The framework of semi-supervised knowledge distillation. As for labeled data information flow, the student model (e.g., ResNet50) is supervised by hard labels (i.e., one-hot vector) and soft labels predicted by the teacher model (e.g., ResNet101-ibn) with cross-entropy loss $\mathcal{L}_c(\theta_s, \tau_{kd})$ and KL loss $\mathcal{L}_{kd}(\theta_s, \tau_{kd})$. As for unlabeled data information flow, the student model is supervised by soft labels predicted by the teacher model with KL loss $\mathcal{L}_{kd}^{u}(\theta_s, \tau_{kd}^{u})$.}
\label{fig4}
\end{figure*}

We collect about 1,000+ minutes video flows from 24 street view videos to generate 301,725 human images and 6-8 samples per subject, we call it YouTube-Human. 
Some examples of YouTube-Human are shown in Fig.~\ref{fig3}. 
Especially, the collected images are captured by the same mobile camera. 
The YouTube-Human dataset mainly has the following properties:
\begin{itemize}
    \item \textbf{Large time span.} The street view videos are collected from diverse seasons where pedestrians wear short sleeves and shorts, as well as long clothes and long pants. In addition, the collected videos were recorded during both daytime and nighttime so that pedestrian images have abundant illumination changes.
    \item \textbf{Different resolutions.} The images are captured at different distances, so the resolutions of person images in YouTube-Human are various. 
    \item \textbf{Occlusion.} Many images are occluded by moving obstacles (e.g., cars, other persons) and static ones (e.g., trees, barriers), resulting in existing occluded person images in YouTube-Human. 
\end{itemize}

To summarize, YouTube-Human has diverse person images, which is beneficial for training a generalizable person re-id model.

\subsection{Semi-supervised Knowledge Distillation}
\subsubsection{Base Model Training} 
In the DG-ReID setting, images from multiple source domain $\mathcal{D}_s=\{(x_i, y_i) \}_{i=1}^{N}$ with $K$ classes are provided, where $N$ is the number of images in $\mathcal{D}_s$ and $y_i$ is the hard label.
We aim to learn a network: $F_{\theta}$: $\mathcal{X} \to \mathcal{R}^{C}$ that composes of a feature extractor $\mathcal{F}$ and a classifier $\mathcal{C}$ for learning feature representations. 
For each training image $x$, its output logits can be represented as $\mathbf{d} = \mathcal{C}_t(\mathcal{F}_t(x))$. 
The predictive probability vector $\mathbf{p}$ can be obtained by a softmax function on the logits, where $\mathbf{p}(k)$ is defined as
\begin{equation}
\begin{array}{l}
\displaystyle \mathbf{p}(k) = \frac{\exp(\mathbf{d}_k/\tau)}{\sum_{j=1}^{K}\exp(\mathbf{d}_j/\tau)},
\end{array}
\end{equation}
where $\tau$ is a temperature hyper-parameter.
As $\tau$ grows, the probability distribution generated by the softmax function becomes smoother, providing more information to learn the whole network. 
Let $y_i\in\{1,\ldots,K\}$ denote the ground truth label of $x_i$, and the cross-entropy loss below is used to optimize the whole network,
\begin{equation}
\begin{array}{l}
\displaystyle \mathcal{L}_c(\theta, \tau_c) = -\sum_{i=i}^{N}\sum_{j=1}^{K}\log\mathbf{p}(y_{i,j}).
\end{array}
\end{equation}

\subsubsection{Knowledge Distillation with Labeled Data}
Knowledge distillation (KD) helps to improve the generalization capability of feature representation models, it means that a small student model could exploit knowledge from a large teacher model.
$\mathcal{F}_s$ and $\mathcal{F}_{t}$ respectively denote a student model and a teacher model, while their classifiers are respectively defined as $\mathcal{C}_s$ and $\mathcal{C}_t$.
Given one labeled data $x$, KD aims to make the predictive probability vector $\mathbf{p}_s = softmax(\mathcal{C}_s(\mathcal{F}_s(x)))$ by a student model approximate probabilities $\mathbf{p}_t = softmax(\mathcal{C}_t(\mathcal{F}_t(x)))$ (i.e., soft labels) predicted by a teacher model, indicating the supervision from distillation. 
We use the Kullback-Leibler (KL) divergence between $\mathbf{p}_s$ and $\mathbf{p}_t$ to train KD in the following, 
\begin{equation}
\begin{array}{l}
\displaystyle \mathcal{L}_{kd}(\theta_s, \tau_{kd}) = \sum_{i=1}^{N} \sum_{j=1}^{K}\mathbf{p}_{t_{i,j}} \log(\frac{\mathbf{p}_{t_{i,j}}}{\mathbf{p}_{s_{i,j}}}).
\end{array}
\end{equation}
Here $n$ is the dimension of classifier, and $\theta_s$ is feature extractor and classifier parameters of the student network.

\subsubsection{Knowledge Distillation with Unlabeled Data}
To effectively utilize unlabeled data, we design an additional student classifier $\mathcal{C}_s^{u}$ whose dimension is the same as that of the teacher classifier. 
Given one unlabeled data $x_u$, its predictive probability vector obtained by the teacher model is presented as $\mathbf{p}_t^{u} = softmax(\mathcal{C}_t(\mathcal{F}_t(x_u)))$. 
Specially, $\mathbf{p}_t^{u}$ could be regarded as pseudo soft labels. 
In the same manner, we obtain the predictive probability vector $\mathbf{p}_s^{u} = softmax(\mathcal{C}_s^{u}(\mathcal{F}_s(x_u)))$ of the student, and then we use the KL divergence below to optimize the student network,
\begin{equation}
\begin{array}{l}
\displaystyle \mathcal{L}_{kd}^{u}(\theta_s, \tau_{kd}^{u}) = \sum_{i=1}^{N} \sum_{j=1}^{K}\mathbf{p}_{t_{i,j}}^{u} \log(\frac{\mathbf{p}_{t_{i,j}}^{u}}{\mathbf{p}_{s_{i,j}}^{u}}).
\end{array}
\end{equation}

\begin{algorithm}[t]
\caption{Semi-supervised Knowledge Distillation (SSKD)}
\label{alg:Framwork}
\begin{algorithmic}[1] 
\REQUIRE
A labeled data $x$ and a unlabeled data $x_u$; 
Initialized student and teacher network parameters $\theta_s$ and $\theta_t$, and the temperature hyper-parameter $\tau_c$, $\tau_{kd}$ and $\tau_{kd}^{u}$.
\ENSURE Parameter $\theta_s$ of the student network. \\ 
\STATE Train the student network with $\mathcal{L}_c(\theta_s, \tau_c)$ and teacher network with $\mathcal{L}_c(\theta_t, \tau_c)$ using SGD optimizer.
\STATE Freeze the teacher network $\theta_t$, and solve $\mathcal{L}_c(\theta_s, \tau_{kd}) + \mathcal{L}_{kd}(\theta_s, \tau_{kd}) + \mathcal{L}_{kd}^{u}(\theta_s, \tau_{kd}^{u})$ to obtain the student parameter $\theta_s$.
\end{algorithmic}
\end{algorithm}

\subsubsection{Training Procedure}
To train a domain-agnostic model, we first use the labeled data to train a student model and a teacher model with the loss function $\mathcal{L}_c(\theta, \tau_c)$.
Then we perform the semi-supervised knowledge distillation algorithm with labeled data and unlabeled data. 
We freeze the network parameters of the teacher model and only train the student model. 
As shown in Fig.~\ref{fig4}, two information flows are involved in the SSKD framework, the first one is the labeled data information flow that is trained with labeled data using $\mathcal{L}_{kd}(\theta_s, \tau_{kd})$, and the second one is the unlabeled data information flow that is trained with unlabeled data using $\mathcal{L}_{kd}^{u}(\theta_s, \tau_{kd}^{u})$. 
Meanwhile, the student model is also supervised by the classification loss function $\mathcal{L}_c(\theta, \tau_c)$ with labeled data. 
Finally, the proposed SSKD is defined as 
\begin{equation}
\begin{array}{l}
\displaystyle \mathcal{L}_{total} = \mathcal{L}_c(\theta_s, \tau_{kd}) + \mathcal{L}_{kd}(\theta_s, \tau_{kd}) + \mathcal{L}_{kd}^{u}(\theta_s, \tau_{kd}^{u})
\end{array}
\end{equation}

\begin{table}[t]
\scriptsize
\centering
\caption{The details of 20 person re-id datasets in \textbf{FastHuman}. These datasets collected from diverse application scenarios make them very suitable for domain generaliable re-id task. Market1501, DukeMTMC-reID, MSMT17, PartialREID, PartialiLIDS, OccludedREID and CrowdREID datasets are regarded as unseen domains for testing and the remaining 17 source domains are used for learning models.}
\vspace{-1em}
\label{tab1}
\begin{tabular}{lcccc}
\toprule
\textbf{Source Domain} &  \#subjects & \#images & \#cameras & collection place \\
\midrule
CUHK03 \cite{li2014deepreid}      & 1,090 & 14,096 & 2 & campus \\ 
SAIVT \cite{bialkowski2012database} & 152   & 7,150  & 8 & buildings \\
AirportALERT \cite{gou2018systematic} & 9,651 & 30,243 & 6 & airport \\
iLIDS \cite{zheng2009associating}       & 300   & 4,515  & 2 & airport \\
PKU  \cite{ma2016orientation}        & 114   & 1,824  & 2 & campus \\
PRAI \cite{ma2016orientation}  & 1,580 & 39,481 & 2 & aerial imagery \\
SenseReID \cite{zhao2017spindle}   & 1,718 & 3,338  & 2 & unknown \\
SYSU \cite{wu2020rgb} & 510   & 30,071 & 4 & campus \\
Thermalworld \cite{kniaz2018thermalgan}& 409   & 8,103  & 1 & unknown \\
3DPeS  \cite{baltieri20113dpes}      & 193   & 1,012  & 1 & outdoor  \\
CAVIARa  \cite{cheng2011custom}    & 72    & 1,220  & 1 & shopping mall \\
VIPeR  \cite{gray2008viewpoint}      & 632   & 1,264  & 2 & unknown \\
Shinpuhkan \cite{kawanishi2014shinpuhkan2014}  & 24    & 4,501  & 8 & unknown \\
WildTrack \cite{chavdarova2018wildtrack}   & 313   & 33,979 & 7 & outdoor \\
cuhk-sysu \cite{xiao2017joint}    & 11,934& 34,574 & 1& street, movie \\
LPW \cite{song2018region}         & 2,731 & 30,678 & 4 & street \\
GRID \cite{loy2013person}         & 1,025 & 1,275 & 8 & underground \\
\midrule
Total       & 31,423& 246,049 & 57 & - \\
\midrule
\textbf{Unseen Domain} &  \#subjects & \#images & \#cameras & collection place  \\
\midrule
Market1501 \cite{zheng2015scalable}   & 1,501  & 32,217 & 6 & campus \\
DukeMTMC \cite{zheng2017unlabeled}     & 1,812  & 36,441 & 8 & campus \\
MSMT17 \cite{wei2018person}      & 4,101  & 126,441& 15& campus \\
PartialREID \cite{zheng2015partial}      & 60  & 600& 6& campus \\
PartialiLIDS \cite{he2018deep}      & 119  & 238 & 2 & airport \\
OccludedREID \cite{zhuo2018}      & 200  & 2,000& 5& campus \\
CrowdREID \cite{he2020guided}  & 845  & 3,257 & 11 & railway station \\ 
\midrule
Total        & 8,638  & 201,184& 49 & - \\
\bottomrule
\end{tabular}
\vspace{-0.5em}
\end{table}

\subsection{A New Benchmark}
Previous works always compare the performance of the proposed algorithm on different training-testing setups.
There does not exist a standard and large-scale benchmark to verify the performance of DG-ReID so that current methods are hard to be fairly evaluated. 
Besides, some works are evaluated on a small-scale unseen target domain with less than 1,000 images so that contingency factors appear in the results.
And the quality of images in these datasets is relatively low, which cannot truly reflect the model's generalization ability in real-world scenarios. 
For these reasons, a unified benchmark of DG-ReID is required to evaluate the performance of different methods, accelerating the deployment of person re-id methods in practical applications.

To build a benchmark for DG-ReID, we collect 21 domain labeled datasets and one unlabeled dataset in Tabel~\ref{tab1}, and then combine them into a large-scale dataset, named \textbf{FastHuman}.
These datasets have diverse images from different application scenarios including campus, airport, shopping mall, street, and railway station.
It contains 447,233 labeled images of 40,061 subjects captured by 82 cameras and 301,725 unlabeled images collected from YouTube street-view videos so that it is suitable for studying DG-ReID. 
\textbf{FastHuman}, as a new benchmark, helps to learn a domain-general model that can generalize well on unseen domains. 
FastHuman is very challenging, as it contains more complex application scenarios and large-scale training, testing datasets.

In the \textbf{FastHuman}, three commonly used large-scale datasets: Market1501~\cite{zheng2015scalable}, DukeMTMC~\cite{zheng2017unlabeled}, MSMT17~\cite{wei2018person} and four occluded datasets: PartialREID, PartialiLIDS, OccludedREID and CrowdREID as unseen target domains are used for evaluate the generalization ability of models under occlusion and crowded scenarios. 
Besides, the remaining 17 labeled datasets and unlabeled dataset YouTube-Human are considered as source domains for training DG-ReID models. 

\section{Experiments}
\subsection{Experiment Settings}

\textbf{Implementation Details.}
All experiments are conducted based on the open-source code FastReID \url{https://github.com/JDAI-CV/fast-reid}. 
We use ResNet50 (R50) and ResNet101-ibn (R101-ibn) pre-trained on ImageNet as the student model and the teacher model, respectively.
We resize the image size to $256\times 128$ and adopt random cropping, random flipping, color jittering as data augmentation methods to improve the generalization ability of models. 
For labeled data, we integrate all source domain datasets into a large-scale hybrid dataset and set batch to 256, including 64 identities and four images per identity. 
For unlabeled data, we adopt random sampling to sample 48 images per batch.
During model training, we empirically set $\tau_c = 1$, $\tau_{kd} = 16$ and $\tau_{kd}^{u} = 6$, respectively. 
We optimize models with the Adam optimizer with 40 epochs and the warmup strategy is used in the first epochs (the warmup factor is set to 0.1). The learning rate is initialized as 7e-4, which is cosine decayed at each iteration until reaches to 7e-7. 
We conduct all the experiments with PyTorch on four P40 GPUs.

\begin{table}[t]
  \centering
  \scriptsize
  \caption{Two evaluation protocols for DG-ReID. Protocol-1 uses the leave-one-out for M+D+C3+MS, it selects one dataset as unseen target domain and the remaining three dataset as labeled source domains. Protocol-2 uses the 17 source domains in \textbf{FastHuman} as labeled data and Market1501, DukeMTMC, MSMT17 as unseen target domains.}
  \vspace{-1em}
  \label{tab2}
    \begin{tabular}{c|c|c|c}
    \shline
   \multirow{2}{*}{Setting} & \multicolumn{2}{c|}{Training Data} & \multirow{2}{*}{Testing Data} \cr \cline{2-3}
                         & Unlabeled Data & Labeled Data     &      \cr \shline
    Protocol-1  & YouTube-Human &\multicolumn{2}{c}{Leave-one-out for M+D+C3+MS} \cr \hline
    Protocol-2  & YouTube-Human & FastHuman & M, D, MS \cr \shline
    \end{tabular}
\end{table}

\begin{table*}[t]
\centering
\footnotesize
\caption{Performance comparison on the Protocol-1.}
\vspace{-2em}
\begin{center}
\begin{tabular}{l|cc|cc|cc|cc}
\shline
\multirow{2}{*}{Methods} & \multicolumn{2}{c|}{MS+D+C$\to$M} & \multicolumn{2}{c|}{MS+M+C$\to$D} & \multicolumn{2}{c|}{M+D+C$\to$MS} & \multicolumn{2}{c}{MS+M+D$\to$C} \cr \cline{2-9} 
& rank-1 & mAP & rank-1 & mAP & rank-1 & mAP & rank-1 & mAP  \cr \hline
QAConv$_{50}$ \cite{liao2020interpretable} (ECCV'20) & 65.70 & 35.60 & 66.10 & 47.10 &  24.30 & 7.50 & 23.50 & 21.00 \cr
M$^{3}$L \cite{zhao2021learning} (CVPR'21) & 74.50 & 48.10 & 69.40 & 50.5 &  33.00 & 12.90 & 30.70 & 29.90\cr
MD-ExCo \cite{yu2021multiple} (Arxiv'21) & 83.20 & 62.20 & 74.10 & 61.50 & 32.70 & 20.90 & 30.90 & 31.30  \cr
RaMoE \cite{dai2021generalizable} (CVPR'21)& 82.00 & 56.50 & 73.60 & 56.90 & 36.60 & 13.50 & 36.60 & 35.50 \cr \hline
MixMatch~\cite{berthelot2019mixmatch} (NeurIPS'19)&79.51&52.50&71.77&52.77&37.11&14.22&34.50&34.00\cr 
FixMatch~\cite{sohn2020fixmatch} (NeurIPS'20)&79.99&54.50&72.04&53.21&39.63&14.96&33.50&33.03\cr \hline
BoT (R50) \cite{luo2019bag} (CVPRW'19)& 78.24 & 51.45 & 71.42 & 51.67 & 35.30 & 14.50 & 36.10 & 34.50\cr 
FastReID-R50 (student)& 80.51 & 54.72 & 74.24 & 56.62 & 41.80 & 16.34 & 43.00 & 41.44\cr 
FastReID-R101-ibn (teacher) & 84.47 & 61.44 & 76.48 & 61.31 & 46.35 & 19.12 & 45.00 & 45.97  \cr \hline
KD (R101-ibn$\to$R50) & 82.57 & 60.02 & 77.51 & 62.25 & 43.27 & 17.40 & 44.71 & 44.00  \cr 
SSKD (R101-ibn$\to$R50) &  86.25 & 65.60 & 77.89 & 63.89 & 47.17 & 20.03 & 45.41 & 45.84 \cr \shline
\end{tabular}
\vspace{-1em}
\label{tab:protocol1}
\end{center}
\end{table*}

\textbf{Evaluation Protocol.}
There exist some evaluation protocols~\cite{dai2021generalizable} \cite{jin2020style} to evaluate the generalization capability of models.
They combine multiple source datasets including Market1501 (M), DukeMTMC (D), CUHK03 (C3), MSMT17 (MS) to train models, and four small datasets including PRID, GRID, VIPeR and iLIDS are used for testing. 
However, the number of images in these four testing sets is very small, which cannot reflect the generalization ability of models. 
Dai \textit{et al.}~\cite{dai2021generalizable} propose a protocol of the leave-one-out setting for commonly used datasets M+D+C3+MS, and they select one dataset as unseen target domain and the remaining three datasets are used for training. 
Based on the protocol, we propose a new protocol where we select one dataset in M+D+C3+MS as an unseen target domain and the remaining three labeled datasets are combined with YouTube-Human as the training set, which is called Protocol-1. 
To verify the effectiveness of SSKD, we also combine \textbf{FastHuman} with YouTube-Human to train models. 
In the two training settings, M, D, MS are used for testing, which is called Protocol-2.

\textbf{Evaluation Metrics.}
We employ the standard metrics in the literature, i.e., the rank-1 accuracy and the mean Average Precision (mAP), to evaluate the performance. 

\subsection{Comparison with the State-of-the-Art Approaches}
\textbf{Results on Protocol-1.} We evaluate our proposed SSKD framework on protocol-1, which is shown in Table~\ref{tab:protocol1}. 
In this experiment, we mainly compare with methods of four categories: generalization ReID algorithms (QAConv$_{50}$ \cite{liao2020interpretable}, M$^{3}$L \cite{zhao2021learning}, MD-ExCo \cite{yu2021multiple} and RaMoE \cite{dai2021generalizable}), semi-supervised learning algorithms (MixMatch~\cite{berthelot2019mixmatch} and Fixmatch~\cite{sohn2020fixmatch}) and strong baselines for ReID task (BoT \cite{luo2019bag} and FastReID~\cite{he2020fastreid}).
 
It is noted that: 1) The gap between KD (R101-ibn$\to$R50 means that R101-ibn distills R50) and SSKD is significant: SSKD increases from 82.57 to 86.25\%, from 77.51\% to 77.89, from 43.27\% to 47.17\% and from 44.71\% to 45.41\% on MS+D+C$\to$M, MS+M+C$\to$D, M+D+C$\to$MS and M+D+MS$\to$C settings, respectively.
It fully demonstrates that SSKD can effectively use the unlabeled data to improve the generalization ability of the model. 
The improvement mainly comes from two aspects: the first aspect benefits from knowledge distillation that can transfer more knowledge from a big model to a small model; the second aspect benefits from the diversity of unlabeled data, where SSKD assigns pseudo soft labels to unlabeled data to improve the performance of the model. 
2) SSKD reopens a new feature space using unlabeled data to increase the discrete ability of the model so that the performance of the student model surpasses the teacher model.
3) Comparing with other state-of-the-art algorithms, SSKD achieves competitive results which respectively outperforms M3L, MD-ExCo, RaMoE, and BoT by 11.75\% and 8.49\%, 3.05\% and 3.79\%, 4.25\% and 4.29\%, 8.01\% and 6.57\% on MS+D+C$\to$M and MS+M+C$\to$D in term of rank-1 accuracy. 
For M+D+C$\to$MS and M+D+MS$\to$C, SSKD surpasses M3L, MD-ExCo, RaMoE, and BoT by at least 6.00\% in terms of rank-1 accuracy. 
Although the semi-supervised learning algorithms of the classification task also consider the influence of unlabeled data, SSKD also surpasses Fixmatch and MixMatch about 6.00\% in terms of rank-1 accuracy. 
The reason for this is that the classification task is dedicated to a close-set problem, whereas the DG-ReID is an open-set task that does not share the category space between the training and testing domains. 
Therefore semi-supervised learning algorithms cannot be applied directly to this task.

\begin{table}[t]
\centering
\scriptsize
\caption{Performance comparison on the Protocol-2.}
\vspace{-2em}
\begin{tabular}{l|c|c|c}
\shline
\multirow{2}{*}{Methods}&  \multicolumn{3}{c}{FastHuman} \cr \cline{2-4} 
&Market1501 &DukeMTMC & MSMT17  \cr
\shline
MeanTeacher~\cite{tarvainen2017mean} & 89.19 (72.59) & 74.82 (57.41) & 54.90 (25.90) \cr 
MixMatch~\cite{berthelot2019mixmatch} & 88.18 (71.11) & 77.02 (59.49) & 52.99 (25.06) \cr 
FixMatch~\cite{sohn2020fixmatch} & 88.12 (71.57) & 74.55 (56.84) & 52.86 (24.82) \cr \hline
FastReID-R50 (student)   & 87.29 (67.05) & 75.99 (59.62) & 52.45 (24.54) \cr 
FastReID-R101-ibn (teacher) & 90.05 (76.56) & 78.64 (63.19) & 59.99 (30.93) \cr \hline
KD	& 87.32 (68.44) & 78.05 (62.19) & 53.30 (26.22)\cr 
SSKD   & 89.54 (75.03) & 79.05 (63.73) & 56.84 (29.43) \cr \hline
Supervised Learning (R50) & 94.40 (86.10) & 79.44 (62.53) & 74.10 (50.20) \cr 
\shline
\end{tabular}
\label{tab:protocol2}
\end{table}

\begin{table*}[t]
\centering
\footnotesize
\caption{Performance comparison on PartialREID, PartialiLIDS, OccludedREID and CrowdREID datasets based on the occluded/partial re-id protocols.}
\vspace{-1em}
\begin{center}
\begin{tabular}{l|cc|cc|cc|cc}
\shline
\multirow{3}{*}{Methods} & \multicolumn{8}{c}{MS+M+D+C} \cr \cline{2-9} 
 & \multicolumn{2}{c}{PartialREID} & \multicolumn{2}{c}{PartialiLIDS} & \multicolumn{2}{c}{OccludedREID} & \multicolumn{2}{c}{CrowdREID}  \cr \cline{2-9} 
 & rank-1 & mAP & rank-1 & mAP& rank-1 & mAP& rank-1 & mAP \cr \hline
MeanTeacher~\cite{tarvainen2017mean} & 82.67 & 80.33 & 80.55 & 85.12 & 91.30 & 88.76 & 54.40 & 55.00 \cr  
MixMatch~\cite{berthelot2019mixmatch} & 76.00 & 74.50 & 75.63 & 82.59 & 87.50 & 83.87 & 57.60 & 59.44 \cr  
FixMatch~\cite{sohn2020fixmatch} & 76.33 & 73.28 & 80.67 & 85.75 & 86.50 & 80.03 & 58.74 & 61.51\cr \hline 
FastReID (R50) & 80.70 & 77.80 & 79.80 & 85.80 & 90.00 & 86.00 & 53.30 & 56.60   \cr
FastReID (R101-ibn) & 81.70 & 82.20 & 81.50 & 86.70 & 91.70 & 87.40 & 59.00 & 63.70   \cr 
KD (R50) & 84.30 & 83.00 & 83.20 & 87.80 & 91.60 & 89.30 & 57.10 & 62.00   \cr \hline
SSKD (R50) & 85.30 & 83.60 & 84.00 & 88.80 & 92.20 & 89.40 & 57.30 & 61.70  \cr 
SSKD + DSR (R50) & 87.30 & 85.30 & 87.40 & 90.70 & 93.30 & 89.70 & 57.50 & 61.80  \cr 
\end{tabular}
\begin{tabular}{l|cc|cc|cc|cc}
\shline
\multirow{3}{*}{Methods} & \multicolumn{8}{c}{FastHuman} \cr \cline{2-9}
& \multicolumn{2}{c}{PartialREID} & \multicolumn{2}{c}{PartialiLIDS} & \multicolumn{2}{c}{OccludedREID} & \multicolumn{2}{c}{CrowdREID}  \cr \cline{2-9} 
 & rank-1 & mAP & rank-1 & mAP& rank-1 & mAP& rank-1 & mAP \cr \hline
MeanTeacher~\cite{tarvainen2017mean} & 78.33 & 76.63 & 86.55 & 90.35 & 92.30 & 88.46 & 55.90 & 58.84 \cr  
MixMatch~\cite{berthelot2019mixmatch} & 76.67 & 75.69 & 84.03 & 88.86 & 92.70 & 88.32 & 60.04 & 62.12 \cr  
FixMatch~\cite{sohn2020fixmatch} & 72.67 & 71.89 & 84.03 & 89.12 & 91.50 & 87.08 & 60.53 & 62.85 \cr \hline
FastReID (R50) & 74.70 & 73.80  & 84.00 & 88.70& 91.20 &86.50 & 51.00 & 53.90  \cr
FastReID (R101-ibn) & 87.70 & 84.40 & 89.10 & 92.00 & 91.50 & 89.30 & 58.20 & 62.70 \cr 
KD (R50) & 81.70 & 80.40 & 85.70 & 90.40 & 92.00 & 87.80 & 51.00 & 54.70 \cr \hline
SSKD (R50) & 86.00 & 84.50 & 87.40 & 91.20 & 93.50 & 90.00 & 57.70 & 61.00 \cr 
SSKD + DSR (R50) & 88.70 & 86.10 & 89.90 & 91.90 & 94.70 & 90.60 & 59.70 & 62.40 \cr \hline
\shline
\end{tabular}
\end{center}
\label{tab:oclu_result}
\end{table*}

\begin{table}[t]
  \centering
  \scriptsize
  \caption{Two evaluation protocols on the Occluded/Partial ReID datasets for DG-ReID. Protocol-3 uses the M+D+C3+MS as labeled training data. Protocol-4 uses the 17 source domains in FastHuman as labeled training data. Protocol-3 and protocol-4 both adopt the YouTube-Human as unlabeled training data and the Occluded/Partial ReID datasets (OccludedREID, CrowdREID, PartialREID, and PartialiLIDS) as the testing data.}
  \vspace{-1em}
  \label{tab6}
    \begin{tabular}{c|c|c|c}
    \shline
    \multirow{2}{*}{Setting} & \multicolumn{2}{c|}{Training Data} & \multirow{2}{*}{Testing Data} \cr \cline{2-3}
    & Unlabeled Data & Labeled Data     &                               \cr\shline
    \multirow{2}{*}{Protocol-3}  & \multirow{2}{*}{YouTube-Human} & \multirow{2}{*}{M+D+C3+MT} & OccludedREID \cr
      &   &  & CrowdREID \cr \cline{1-3}
    \multirow{2}{*}{Protocol-4}  & \multirow{2}{*}{YouTube-Human} & \multirow{2}{*}{FastHuman} & PartialREID,\cr
      &  &  & PartialiLIDS\cr\shline
    \end{tabular}
\end{table}
\begin{table}[t]
\centering
\footnotesize
\caption{Details of four Occluded/Partial ReID datasets.}
  \vspace{-1em}
\label{tab7}
    \begin{tabular}{l|c|c}
    \shline
    \multirow{2}{*}{Testing Dataset \qquad} &\multicolumn{2}{c}{\#id/\#imgs} \cr \cline{2-3}
     &Gallery & Probe\cr \hline
    PartialREID   &60/300&60/300\cr
    PartialiLIDS  &119/238&119/238\cr
    OccludedREID  &200/1,000&200/1,000\cr
    CrowdREID   &845/2,412&605/835\cr\shline
    \end{tabular}
\end{table}
\textbf{Results on Protocol-2.} 
We evaluate our proposed SSKD on \emph{FastHuman}, which is shown in Table~\ref{tab:protocol2}. Since \emph{FastHuman} contains more than 200K+ labeled training data, SGD optimizer is a better choice to train models for obtaining a better baseline.
We provide both results based on R50 and R101-ibn. All models are evaluated individually on Market1501, DukeMTMC, MSMT17 datasets. 
Besides, we also provide supervised learning (SL) algorithm based on R50 on Market1501, DukeMTMC, MSMT17 datasets. 
SL follows the standard training-testing split protocols~\cite{zheng2015scalable} where 751 subjects with 12,936 images, 750 subjects with 15,913 and 1,041 subjects with 32,621 images are used for training, the remaining images are used for testing.

It is noted that: 1) Benefiting from diverse training data, the performance of SSKD that trained with \emph{FastHuman} close to SL (training data and testing data come from the same domain. 
2) On the one hand, semi-supervised knowledge distillation (SSKD) is a semi-supervised learning framework that can effectively combine labeled data with unlabeled data to improve model generalization ability, it increases the rank-1 accuracy by 1.80\%, 0.90\%, 2,95\% on Market1501, DukeMTMC and MSMT17 compared to KD. On the other hand, SSKD adopts knowledge distillation to enhance the discriminative of the model.
3) The gap between SSKD and SL is significant, SSKD can only alleviate domain bias to some extent, but it still cannot simulate the source domain data distribution, resulting in performance degradation.

\subsection{Evaluation on Occluded Person Datasets}
\subsubsection{Datasets}
\noindent\textbf{PartialREID}~\cite{zheng2015partial} is a specially designed partial person dataset that includes 600 images from 60 people, with 5 full-body images and 5 partial images per person. 
These images are collected at a university campus from different viewpoints, backgrounds, and different types of severe occlusions.

\noindent\textbf{PartialiLIDS}~\cite{he2018deep} is a simulated partial person dataset based on iLIDS~\cite{zheng2009associating}. 
It contains a total of 238 images of 119 people captured by multiple non-overlapping cameras. 
Some images in the dataset contain people occluded by other individuals or luggage. 

\noindent\textbf{CrowdREID}~\cite{he2020guided} is a newly created, which is specifically designed for person Re-ID in crowded scenes. 
The dataset is very challenging because most of the person images in these images to be identified are occluded by the partial body of other persons. 
In this dataset, 2,412 images of 835 identities are used as the gallery set and 845 images of 605 identities are used as the gallery set.

\noindent\textbf{OccludedREID}~\cite{zhuo2018} is an occluded person dataset captured by mobile cameras, consisting of 2,000 images of 200 occluded persons. 
Each identity has 5 full-body person images and 5 occluded person images with different viewpoints, backgrounds, and different types of severe occlusions.

We provide two evaluation protocols for the training dataset and the testing dataset. 
The first protocol uses labeled data: M+D+MS+C and unlabeled data: YouTube-Human as a training dataset. 
The second protocol uses labeled data: source domain in \emph{FastHuman} and unlabeled data: YouTube-Human as the training dataset. 
Four occluded datasets are used for testing the performance of models. Table~\ref{tab7} shows the details of testing datasets.

\subsubsection{Evalutaion Results} 
For comparison on four occluded/partial datasets, we provide a baseline based on FastReID, where ResNet50 (R50) and ResNet101-ibn (R101-ibn) backbones are used in models. 
For the knowledge distillation (KD) algorithm, we still use R101-ibn to distill R50, and R50 is used as the final inference model. 
The semi-supervised knowledge distillation (SSKD) framework is respectively trained with labeled data (e.g., MS+M+D+C or \emph{FastHuman}) and unlabeled data YouTube-Human. 
Besides, we also provide another baseline algorithm Bag-of-Tricks (BoT), and an occluded re-id algorithm named Deep Spatial feature Reconstruction (DSR) for occluded/partial person matching.

\noindent\textbf{Results on Protocol-3.} 
We evaluate our SSKD on Protocol-3, which is shown in Table~\ref{tab:oclu_result} 
The baseline R50 produces a relatively low accuracy: rank-1 accuracy are 80.7\%, 79.8\%, 90.0\% and 53.3\% on PartialREID, PartialiLIDS, OccludedREID and CrowdREID datasets, respectively. 
KD also achieves competitive results on these datasets. 
SSKD performs better than KD, the reasons mainly have: 1) YouTube-Human dataset contains many occluded/partial images as shown in Fig.~\ref{fig3} so that the model has a strong ability to address occlusion cases in the four datasets.
2) SSKD provides a feasible solution for utilizing the large-scale unlabeled data YouTube-Human to learn representations, which effectively promotes the discrimination ability of models. 
Besides, SSKD+DSR achieves the best performance because DSR effectively uses the correlation of spatial features via sparse coding reconstruction.
\begin{figure}[t]
\centering
\includegraphics[height=3.8cm]{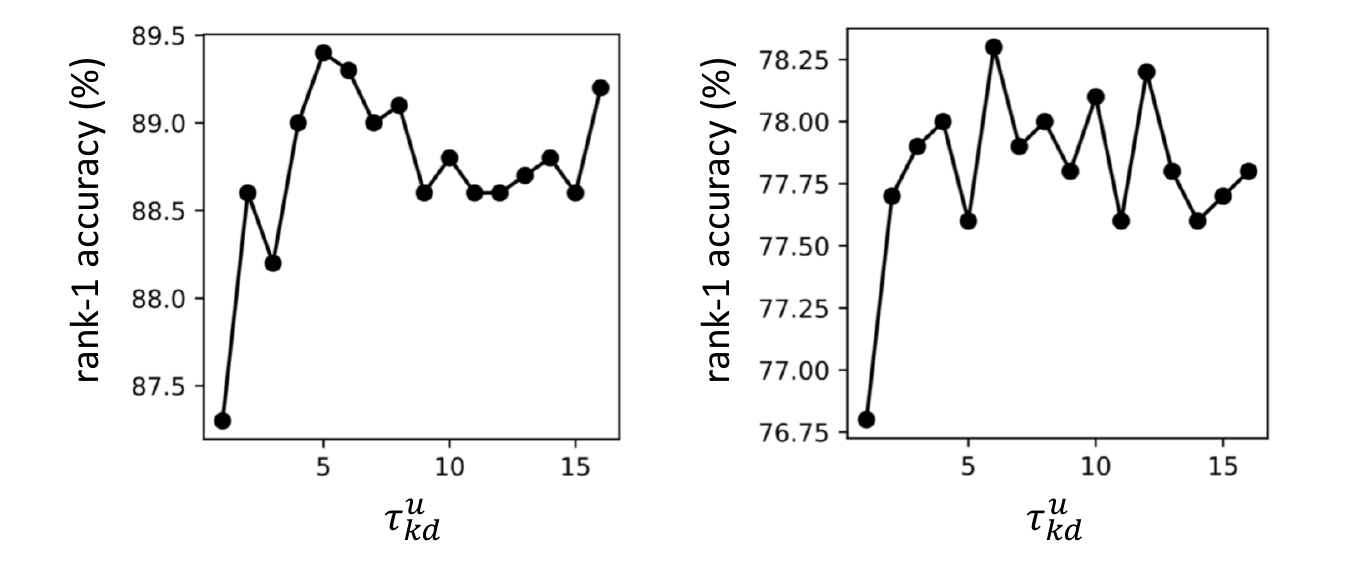}
\vspace{-1.5em}
\caption{Rank-1 accuracy as a function of various $\tau_{kd}^{u}$ on Market1501 and DukeMTMC datasets, respectively.}
\label{fig5}
\end{figure}
\begin{figure}[t]
\centering
\includegraphics[height=10cm]{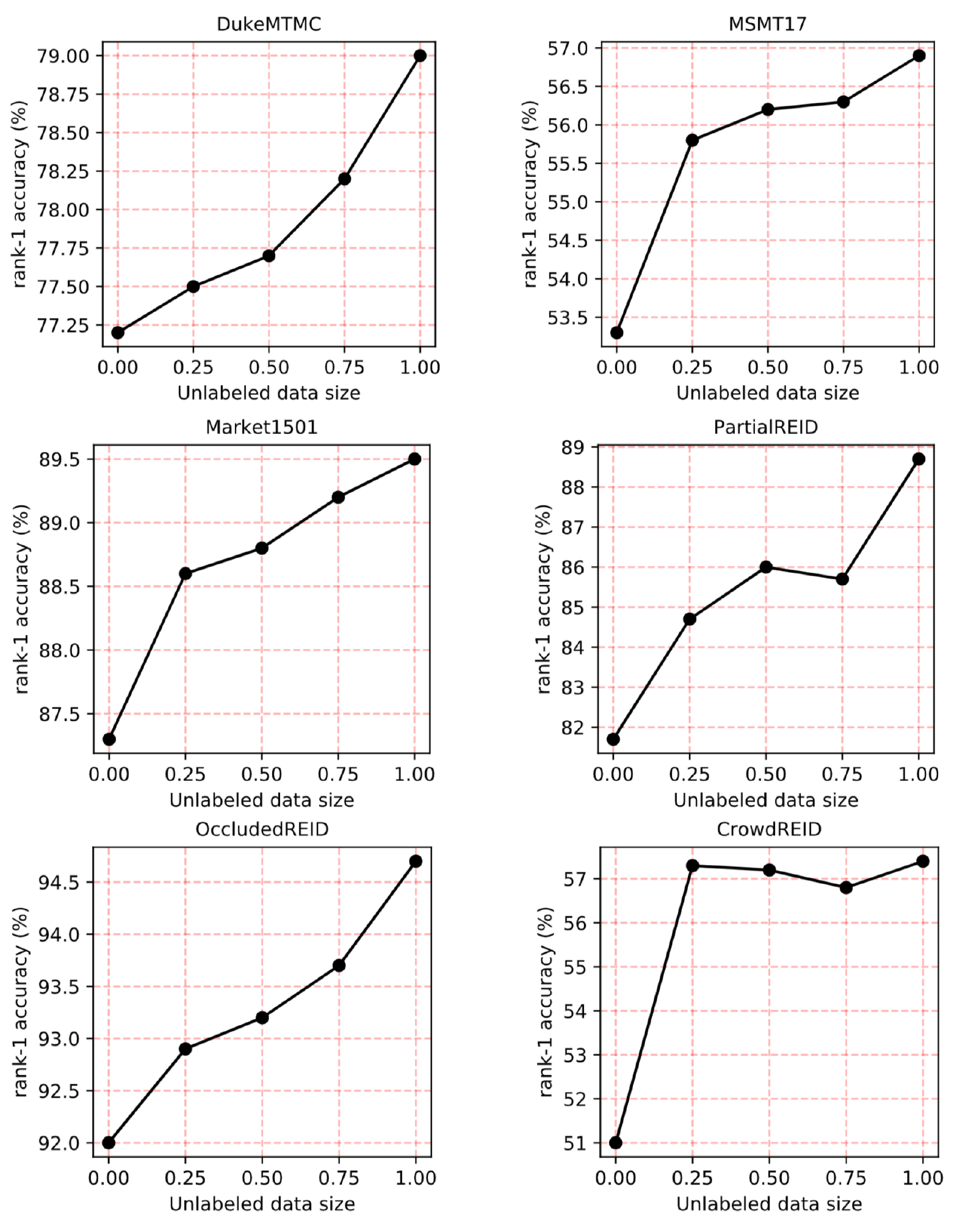}

\caption{Rank-1 accuracy as a function of the different data size on Market1501, DukeMTMC, MSMT17, PartialREID, OccludedREID, CrowdREID datasets, respectively.}
\vspace{-1em}
\label{fig:datasize}
\end{figure}
\noindent\textbf{Results on Protocol-4.} 
We also evaluate our SSKD on Protocol-4, which is shown in Table~\ref{tab:oclu_result}. It is noted that: 1) We can find that SSKD performs much better than KD especially on PartialREID and CrowdREID, it increases from 81.70\% to 86.00\% and from 51.00\% to 57.70\%, respectively; 2) Benefiting from deep spatial reconstruction, SSKD+DSR outperforms SSKD by 2.70\%, 2.50\%, 1.20\%, and 2.00\% at the rank-1 accuracy on the four datasets, respectively; 3) The knowledge distillation indeed improves generalization ability of models by transferring knowledge from a big model to a small model. In conclusion, the proposed SSKD framework cleverly infiltrates unlabeled training data into knowledge distillation, which greatly improves the generalization performance of models. 

\subsection{Ablation Study}
\noindent\textbf{Influence of the temperature of unlabeled data.} We investigate the influence of temperature $\tau_{kd}^{u}$ in $\mathcal{L}_{kd}^{u}$ with unlabeled data. 
We respectively set $\tau_{kd}^{u}$ from 1 to 16. 
Due to the space limit, we only provide the results of SSKD that uses \emph{FastHuman} as training data. 
Fig.~\ref{fig5} shows rank-1 accuracy curves as function of $\tau_{kd}^{u}$.
We find that SSKD achieves the best performance when we set $\tau_{kd}^{u} = 6$, it respectively achieves 89.3\% and 78.4\% rank-1 accuracy. 
In addition, we can also observe from Fig.~\ref{fig5} that $\tau_{kd}^{u}$ is not sensitive to experimental results, it can obtain relatively good experimental results when we set $\tau_{kd}^{u}$ to be in the range from 2 to 16.

\noindent\textbf{Influence of the Size of Unlabeled Data.} We investigate the influence of unlabeled data size for the generalization ability of models. 
We respectively conduct experiments that use 0, 1/4, 1/2, 3/4, 4/4 of all unlabeled data and evaluate on multiple data. 
Fig.~\ref{fig:datasize} shows the rank-1 accuracy curves as a function of unlabeled data size. 
We can observe that the rank-1 accuracy increases with the growth of unlabeled data size. 
Such experimental results indicate that the scale of unlabeled data has a crucial influence on the generalization ability of the model.
\begin{table}[t]
\centering
\footnotesize

\caption{Performance comparison using different backbone as the teacher model. Note: the student model is R50.}
\vspace{-1em}
\begin{center}
\begin{tabular}{l|c|c|c|c}
\shline
Teacher Model & R50 & R101-ibn & ResNest200 & ResNest269 \cr \hline
DukeMTMC & 77.96 & 78.32 & 78.28 & 79.22    \cr
MSMT17 & 55.86 & 56.29 & 57.34 & 57.35      \cr 
Market1501 & 89.99 & 89.31 & 90.29 & 90.86  \cr 
PartialREID & 79.67 & 85.33 & 84.67 & 84.00 \cr 
PatialiLIDS & 84.87 & 88.24 & 86.55 & 87.32 \cr 
OccludedReID & 93.90 & 92.60 & 93.60 & 93.60 \cr 
CrowdReID & 54.68 & 57.39 & 58.34 & 59.24   \cr \shline

\end{tabular}
\label{tab:different teacher}
\end{center}
\end{table}

\noindent\textbf{Benefits of Strong Teacher Model.} In general, the stronger the teacher network, the higher the performance of the student network. 
To investigate the influence of the teacher model, we select R50 as the student network, and R50, R101-ibn, ResNest200, ResNest269 as the teacher network respectively. Table~\ref{tab:different teacher} shows the rank-1 accuracy of using different teacher networks. ResNet-269 as the teacher network can achieve the best performance, it outperforms R50 (distills itself) by 1.28\%, 1.49\%, 0.87\%, 4.33\%, 2.45\%, -0.3\% and 4.56\% at rank-1 accuracy on the seven datasets, respectively. 
Such results suggest that stronger models help to improve the generalization ability of the student model without affecting the inference speed of the model.

\section{Conclusions}
We have proposed a novel framework called Semi-supervised Knowledge Distillation (SSKD) to generalizable person re-identification. The proposed method provides a feasible solution that combines labeled data with unlabeled data to learn domain-general features by knowledge distillation. More importantly, we provide a large-scale unlabeled person dataset named YouTube-Human that contributes to providing diverse data for improving the generalization ability of models. To promote generalizable re-id forward, we also introduce a new benchmark that includes multi-domain data and shows more challenges for this task.

\ifCLASSOPTIONcaptionsoff
  \newpage
\fi

{\small
\bibliographystyle{ieee_fullname}
\bibliography{egbib}
}




\end{document}